\pdfoutput=1
\documentclass{ngsm2024} 

\usepackage{times}
\usepackage{latexsym}
\usepackage{subcaption}
\usepackage{svg}
\usepackage{bbding}

\usepackage{microtype}

\usepackage[capitalise]{cleveref}

\usepackage{amsmath}

\title{Associative Recurrent Memory Transformer}

\optauthor{%
\Name{Ivan Rodkin}$^1$ \quad
\Name{Yuri Kuratov}$^{2,1}$ \quad
\Name{Aydar Bulatov}$^1$ \quad
\Name{Mikhail Burtsev}$^3$ \\ \\
\normalfont{$^1$Neural Networks and Deep Learning Lab, MIPT, Dolgoprudny, Russia \\
$^2$AIRI, Moscow, Russia \quad $^3$London Institute for Mathematical Sciences, London, UK \\
\texttt{\{rodkin.id,yurii.kuratov,bulatov.as\}@phystech.edu, mb@lims.ac.uk}
}
}

\begin{document}
\maketitle
\begin{abstract}
This paper addresses the challenge of creating a neural architecture for very long sequences that requires constant time for processing new information at each time step. Our approach, Associative Recurrent Memory Transformer (ARMT), is based on transformer self-attention for local context and segment-level recurrence for storage of task specific information distributed over a long context. 
We demonstrate that ARMT outperfors existing alternatives in associative retrieval tasks and sets a new performance record in the recent BABILong multi-task long-context benchmark by answering single-fact questions over 50 million tokens with an accuracy of 79.9\%. The source code for training and evaluation is available on \href{https://github.com/RodkinIvan/associative-recurrent-memory-transformer}{github}.

\end{abstract}

\section{Introduction}

Memory plays a crucial role in creating models capable of processing extremely long contexts and utilizing remote past information.
Starting from RNNs and evolving through LSTM~\citep{lstm_1999hochreiter} and Memory Networks~\citep{weston2014memory,sukhbaatar2015endtoend}, we are now in the era of Transformer-based~\citep{vaswani2017attention} Large Language Models~\citep{brown2020languagegpt,touvron2023llama,openai2024gpt4}. Various methods for extending transformers context length have emerged~\citep{peng2023yarn,chen2023longlora,zhang2024soaring}, including approaches based on transformer segment-level recurrence~\citep{dai2019transformerxl,raecompressive2019,bulatov2022recurrent,chevalier2023adapting} and novel architectures that combine the efficiency of transformer parallelization during training with recurrence at inference~\citep{gu2021s4,peng2023rwkv, fu2023hungry,gu2023mamba,de2024griffin}. Alternatively, Retrieval-Augmented Generation (RAG) focuses on retrieving information from external storage~\citep{realm_guu2020, retro_borgeaud2022, replug_shi2023} or self-retrieving from past inputs~\citep{wu2022memorizing,rubin2023long}. However, retrieval fails on complex tasks that require reasoning over multiple pieces of information~\citep{kuratov2024search}.

In this work we propose the Associative Recurrent Memory Transformer (ARMT) as an extension of the segment-level recurrent model RMT~\citep{bulatov2022recurrent} with associative memory. Compared to RWKV~\citep{peng2023rwkv} and Mamba~\citep{gu2023mamba}, which use association-based techniques, ARMT benefits from full local self-attention and has constant time and space complexity of processing new segment, similar to RMT. To study ARMT performance we use the BABILong~\citep{kuratov2024search} benchmark, because it allows to generate test samples up to 50 million tokens and beyond, compared to other methods. Additionally, we use Associative Retrieval task with multiple key-value pairs to estimate memory capacity of models.

Main contributions of this work include: 
\textbf{(1)} a novel ARMT architecture for long context processing with segment-level recurrence and associative memory;
\textbf{(2)} demonstration that ARMT outcompetes existing memory based models like RMT~\citep{bulatov2022recurrent} and Mamba~\citep{gu2023mamba} on associative retrieval and long context processing tasks, achieving 80\% accuracy of single fact QA on unprecedented input size of 50 million tokens;
\textbf{(3)} an original method to evaluate memory capacity in associative retrieval task. 

\begin{figure*}
\centering
\vspace{-15pt}
\includegraphics[width=\textwidth]{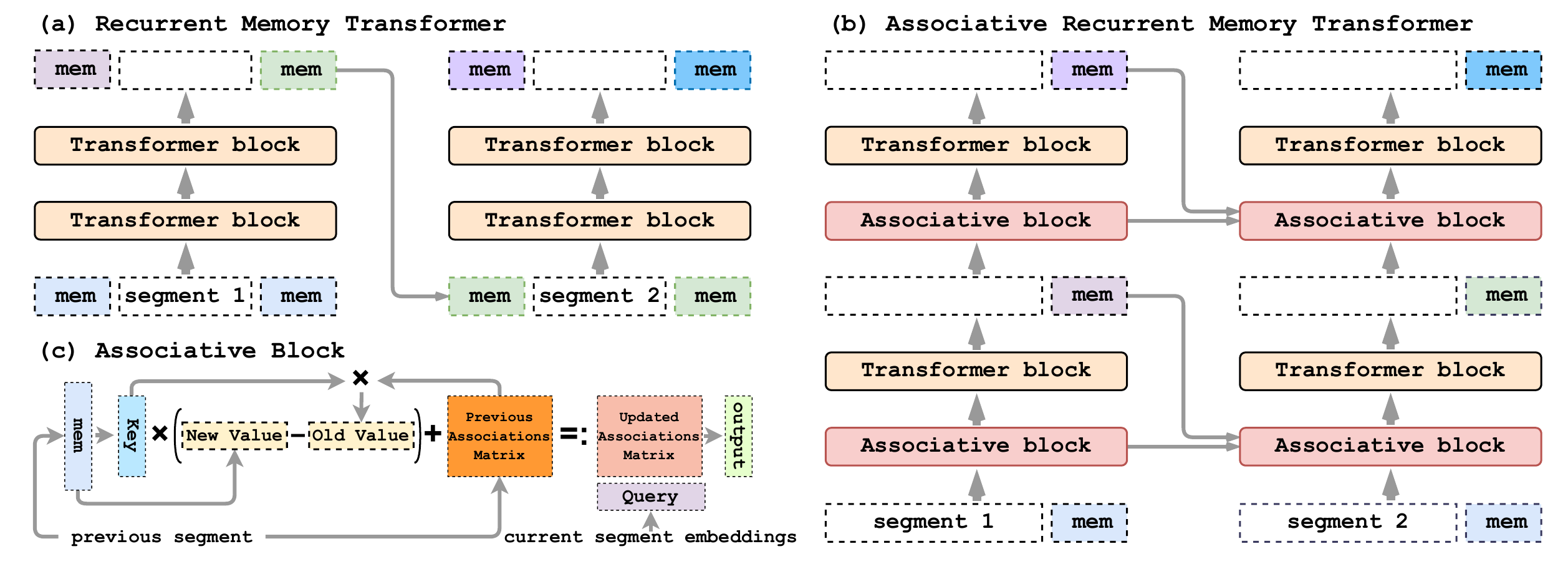}
\caption{\small{\textbf{ARMT augments the transformer's layers with associative memory.} (a) RMT architecture. (b) ARMT adds associative memory processing to each layer. (c) Associative memory is updated with layerwise memory representations.}}
\label{fig:rmt_armt}
\vspace{-10pt}
\end{figure*}

\section{Associative Recurrent Memory Transformer}
\label{sec:armt}
\setlength{\belowdisplayskip}{3pt} 
\setlength{\belowdisplayshortskip}{3pt}
\setlength{\abovedisplayskip}{3pt} 
\setlength{\abovedisplayshortskip}{3pt}

We extend RMT~\citep{bulatov2022recurrent}(\cref{fig:rmt_armt}a) by addition of layerwise associative memory $A_s^l$ over segmented input $X_{s}^l$ (\cref{fig:rmt_armt}b). At every input segment $s$ for each layer $l$ memory tokens $M_{s-1}^{l+1}$ generated for preceding segment are added to $A_{s}^l$ (\cref{fig:rmt_armt}c) used to update input sequence and memory embeddings:
\begin{gather*}
[X^{l+1}_s; M^{l+1}_s] = TrBlock(AssocBlock([X_{s}^l; M_{s}^l], A_s^l)); \quad 
A_s^l = MemUpdate(A_{s-1}^{l}; M_{s-1}^{l+1}).
\end{gather*}

The mechanism of associative block (\cref{fig:rmt_armt}c) is similar to linear transformers~\citep{katharopoulos2020transformers}, but attends only to special memory tokens and is calculated differenty. After each segment, memory tokens are converted to keys and values via linear mapping and then stored in quasi-linear key-value memory \citep{schlag2021linear} using non-linearity $\phi$. Given a memory token $m_i \in M^{l+1}_s$, we calculate the keys, values, and importance scalars $\beta_i$. We then recall the previous association $\overline{v}_i$ with this key, add the new value $v_i$ to the memory, erase the previous value $\overline{v}_i$ associated with $k_i$, and update the normalization vector.
\begin{gather}
k_{i},v_{i} =W_K m_{i},W_V m_{i}; \quad \beta_{i} = \sigma(W_\beta m_i); \quad
A_0^l = \vec{0}; \quad  z_0^l = \vec{0};\\
\overline{v}_i = \frac{A_{s-1}^l \phi(k_i)}{(z_{s-1})^T \phi(k_i)}; \quad \gamma_i = 1 - \frac{(z_{s-1})^T\phi(k_i)}{\|\phi(k_i)\|^2}; \\
A_s^l = A_{s-1}^l + \sum_i \beta_i (v_i - \overline{v}_i) \otimes \phi(k_i); \quad z^l_{s} = z^l_{s-1} + \sum_i\gamma_i \phi (k_i).
\end{gather}
Once we updated $A_s^l$ with information from previous segment, we recall an association $y_j$ for a token $x_j$. Associations $y_j$ for each token in the segment are then passed to the next transformer layer:
 \begin{align}
q_j = W_Q x_j; \quad 
y_j = \frac{A_s^l \phi(q_j)}{(z^l_s)^T \phi(q_j)}.
\end{align}

For the non-linearity function $\phi$, we used the proposed in \citep{schlag2021linear} DPFP-3 function because, in this paper, it has shown significant superiority over other methods, which is also consistent with our findings.

Note that without $\gamma_i$ ($\gamma_i = 1$) this approach suffers from catastrophic forgetting on some tasks. The reason is that while we erase the information $\overline{v}_i$ from the $A$-matrix, the corresponding keys remain in the normalization vector $z_s$. As shown in our experiments (\cref{fig:rewrite_ablation}), this problem becomes significant when performing hundreds of erase-insert operations with associative memory. To overcome this, we propose to take into account the previous keys in $z_s$ when updating it (details are in \cref{sec:gamma_appx}).


To determine which part contributes the most to ARMT performance, we also studied RMT with layerwise recurrent memory without associative memory block (Parallel Memory RMT, or PRMT; see \cref{fig:prmt_fig} in  \cref{sec:prmt}).



\section{Evaluation of Associative Retrieval and Long Context Memory Retention}

We test memory capacity of ARMT in comparison to recent computationally efficient long-context models Mamba~\citep{gu2023mamba} and RMT~\citep{bulatov2022recurrent} 
on the following two variants of associative retrieval.\footnote{We did our best but failed to train RWKV model for associative retrieval and BABILong benchmarks, \cref{app:rwkv}.}





\begin{figure}
\vspace{-15pt}
\includegraphics[width=\textwidth]{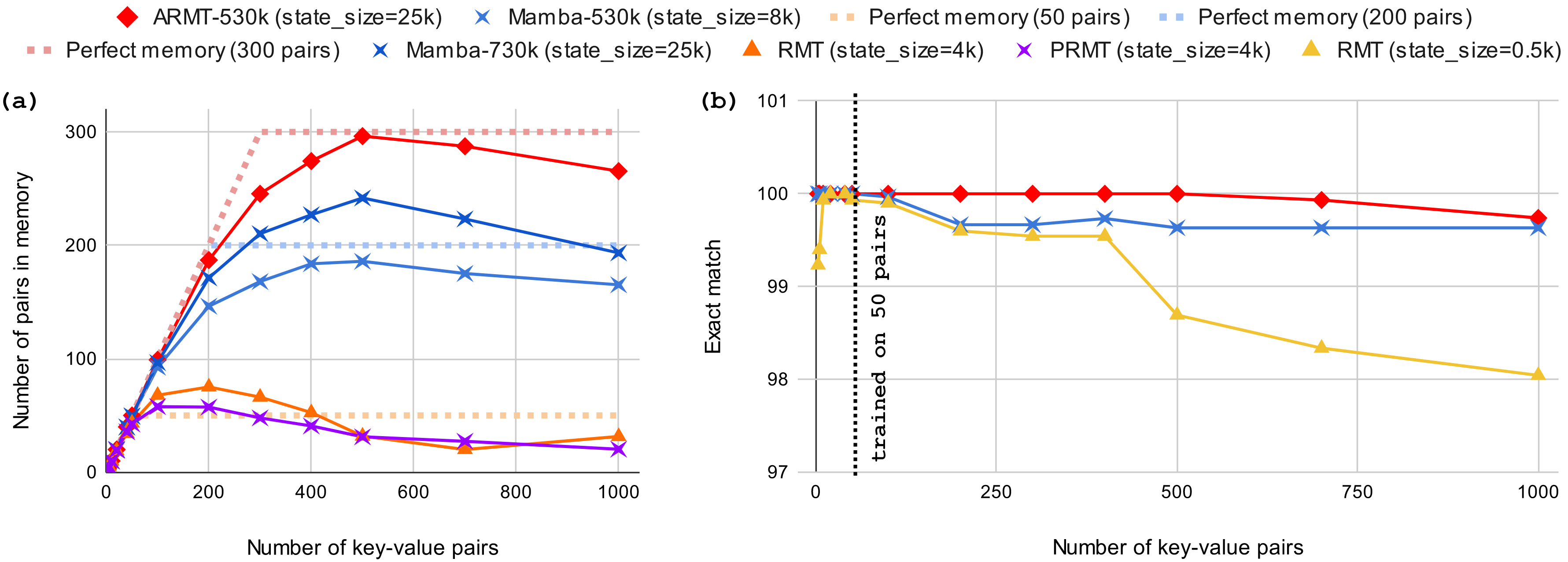}
\vspace{-15pt}
\caption{\small{\textbf{ARMT demonstrates strong performance on associative memory tasks.} \textbf{(a)} The estimated number of pairs, stored in memory after processing the context with key-value pairs. \textbf{(b)} ARMT is more accurate at operations in memory. Being trained only on 50 key-value pairs from Associative Retrieval Rewrite task, ARMT performs accurate even on 500 memory updates. So the observed generalization factor is 10 (500 pairs / 50 pairs). All data are averaged over 3 runs except RMT and PRMT with 2 runs. }}
\label{fig:AR}
\vspace{-10pt}
\end{figure}

\textbf{Remember task} requires memorization of \textit{all} key-value pairs with unique keys from the prior context with subsequent recalling a value corresponding to one of the keys (\cref{fig:AR}a). We estimate the total number of key-value pairs stored in memory, based on the exact-match metric (details are in \cref{sec:mem_cap_appx}). Since ARMT has the same recurrent memory size as Mamba, calculated as the number of floats in recurrent states, it can be concluded that ARMT makes better use of its internal associative memory. Both ARMT and Mamba outperform RMT on this task.
PRMT does not improve RMT performance (\cref{fig:AR}a). This indicates that the associative memory plays a critical role in ARMT performance compared to RMT. Additionally, we ablated ARMT on normalization correction, as detailed in \cref{app:ablation}.

In \textbf{Rewrite task} the keys are not unique and the goal is to recall the \textit{latest} value that corresponds to one of the keys from the prior context. This task evaluates the model's ability to dynamically change the memory storage. The results, shown in \cref{fig:AR}b, indicate that ARMT is robust to the number of memory rewrite operations, while RMT and Mamba experience slight degradation after exceeding their training lengths. Notably, ARMT maintains perfect memory recall on lengths exceeding 10 times those used in training.

\begin{figure*}
    \centering
    \vspace{-15pt}
    \includegraphics[width=\textwidth]
    {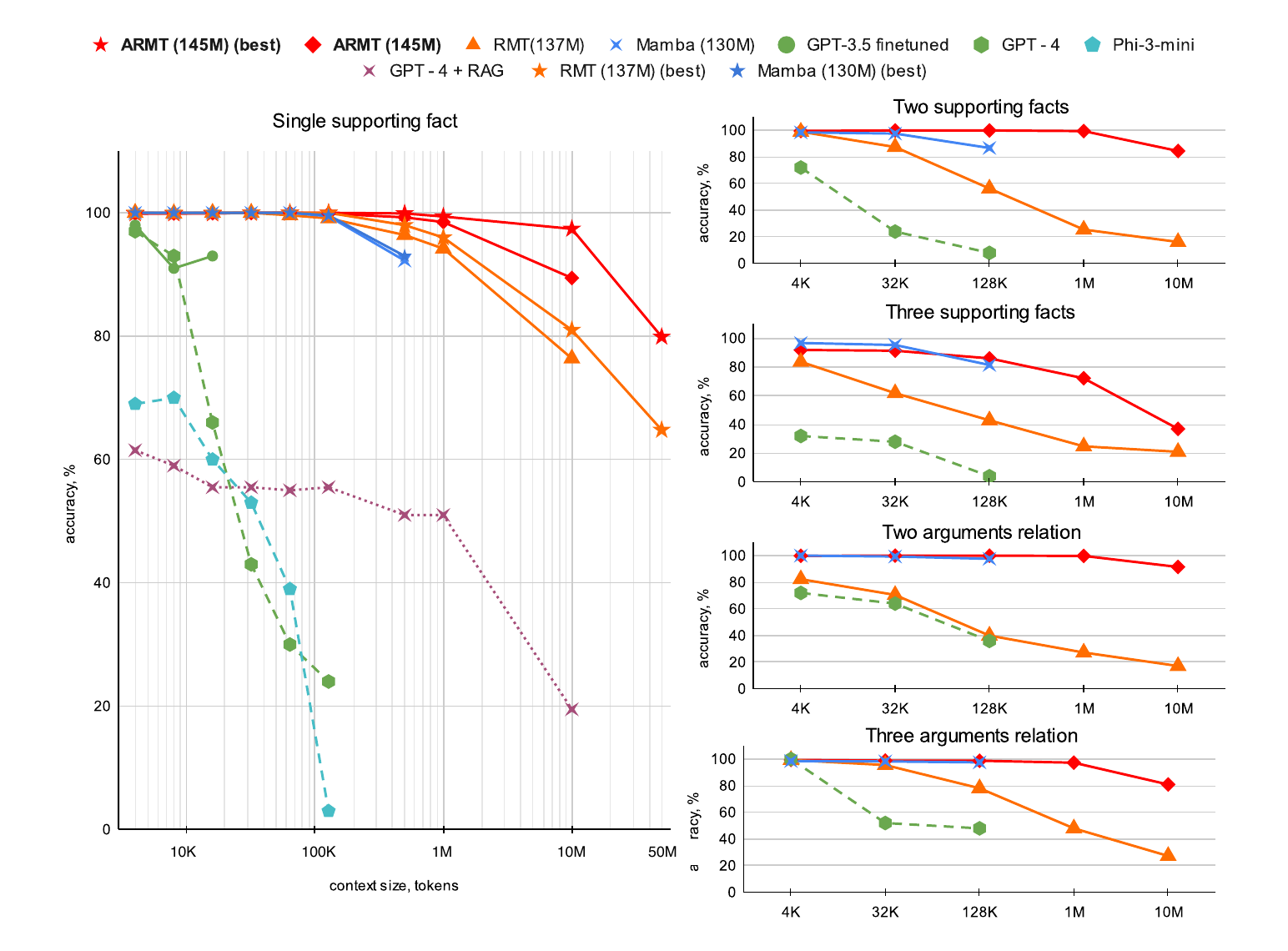}
    \vspace{-15pt}
    \caption{\small{\textbf{ARMT sets a record in long-context processing with reasonable performance on 50 million tokens.} Accuracy of models on different lengths from Babilong benchmark: panels a-e represent QA1-5 tasks.}}
    \label{fig:babilong}
    \vspace{-10pt}
\end{figure*}

We augment the GPT-2 (137M) model with ARMT to solve the BABILong tasks from a recently introduced benchmark for long context processing in a question-answering form~\citep{kuratov2024search}.
To answer BABILong questions correctly, models have to find multiple relevant facts distributed across long natural contexts with distractor facts, and combine information from relevant facts. We use the exact match metric to evaluate models' performance.
As shown in \cref{fig:babilong}, ARMT outperforms the competitors in the majority of tasks, especially on long sequences. Being trained on 16k tokens only, it strongly  performs up to 50 million tokens on QA1 single supporting fact (\cref{fig:babilong}a) and up to 10 million tokens on more complex tasks requiring multi-hop reasoning (\cref{fig:babilong}b-e). We observe 60x length-generalization on these tasks (1M / 16k), while Mamba has 8x length-generalization (128k / 16k). For Mamba-130m we consider only the lengths up to 128k due to its implementation limitations (see \cref{sec:mamba_slow}, and \cref{sec:babilong_appx,sec:curriculum} for training details).
 
\section{Conclusion}

In this work, we propose and evaluate recurrent memory transformer augmented with associative memory mechanism for long-context processing and find that it scales up to an unprecedented 50 million tokens on the BABILong benchmark.
ARMT architecture adds an associative memory mechanism for segment-level recurrent model RMT. Based on our evaluation on associative retrieval tasks ARMT demonstrates significant advantage in memory capacity and generalisation compared to Mamba and RMT. On  BABILong benchmark ARMT dominates alternatives on medium sizes up to 500K tokens and the only approach with high performance across all five tasks in the range of 500K-10M tokens.

We conclude that ARMT holds great promise for long-range tasks because of its improved memory capacity, its ability to efficiently handle large numbers of rewrite operations with memory, its ability to extract only relevant information from memory during inference, and its generalization to much longer sequences than it was trained on.
We also assume that the ARMT can be used for language modeling (\cref{sec:lm_appx,sec:limit}). Despite the current results, we believe there is potential to enhance its performance on LM task through further research and optimization. 

Since all of the results in this study are obtained on relatively small (137M) models, we also assume that the scaling of our methodology and its combination with other techniques can reveal the significant potential for modern large language models. We believe that investigating the properties of recurrent associative memory remains an exciting area of research. 

\section*{Acknowledgements}
We are thankful to SberDevices for granting us access to additional computational resources.
This work was supported by a grant for research centers, provided by the Analytical Center for the Government of the Russian Federation in accordance with the subsidy agreement (agreement identifier 000000D730324P540002) and the agreement with the Moscow Institute of Physics and Technology dated November 1, 2021 No. 70-2021-00138.


\bibliography{custom}

\clearpage
\appendix 
\section{Related Work}

AutoCompressor~\citep{chevalier2023adapting} is a strategy that stacks memory to minimize information loss at the price of quadratic computation cost.

\textbf{Recurrent Memory Transformers} \quad Recent challenges in long-context processing tasks demonstrated recurrent memory superiority over attention mechanism \citep{kuratov2024search, zhang2024soaring} (\cref{fig:rmt_armt} (a)). It was shown that this type of memory performs well even in contexts of size 11M \citep{kuratov2024search}. But still, this memory has some issues with capacity and training. Capacity remains limited as the suggested memory states are limited to a small number of memory tokens. The training is still challenging, as the whole training process requires backpropagation through time for hundreds of layers. Our approach is supposed to mitigate these problems by leveraging the association matrix as a connector for different segments. In contrast to RMT, it has different parameters for memory (linear attention projections described in \cref{sec:armt}) and makes this memory hierarchical by creating different association matrices for different layers.

\textbf{Context explosion prevention} \quad In the attention sinks paper \citep{xiao2023efficient}, authors demonstrated the need for some sinking tokens for attention for efficient extrapolation in long contexts. Recurrent Memory in RMT \citep{bulatov2022recurrent} as well as our model successfully perform this function. In RMT the memory tokens can act as attention sinks while in our model the very association matrix can play this role.

The ARMT can also be thought of as a kind of compressed-memory RMT~\citep{bulatov2022recurrent}, that attends to all previous memory tokens with layerwise memory.

\textbf{Recurrent Sequence Models} \quad The problem of the quadratic cost of attention mechanism led to the development of recurrent architectures with transformer-like performance \citep{de2024griffin}. The vast majority of them are at least related to the State-Space Models (SSMs) \citep{gu2023mamba, fu2023hungry, peng2023rwkv, sun2023retentive}. Despite comparable performance with transformers on LM tasks, SSMs are known to be less efficient in memorization tasks, especially when the question is asked after the information \citep{jelassi2024repeat}. Our model performs well even on these types of tasks, because it has the large and flexible storage for keeping the associations in memory, simultaneously having the direct access to the local context via the vanilla attention.

\section{Memory capacity estimation}
\label{sec:mem_cap_appx}
\textbf{Theorem}:

Given:

$$
\text{exact\_match} = \alpha; \quad n = \text{number of pairs}; \quad v = \text{number of possible values}
$$

Then the number of memorized pairs can be estimated with the formula:

$$
k = \frac{nv\alpha - n}{v-1}
$$
\textbf{Proof}:

We can precisely predict the associated value if we remember k pairs and then extract the key from these pairs. We output the random value if we obtain the key from any other pair. As a result, the following is the mathematical expectation of an exact match:

$$
\alpha = \frac{k}{n}\cdot 1 + \frac{n-k}{n}\cdot \frac{1}{v} = \frac{k}{n}(1- \frac{1}{v}) + \frac{1}{v} = \frac{k(v-1)+n}{nv}
$$

$$
k = \frac{nv\alpha - n}{v-1} 
$$

\textbf{Additionally}:

$$
k = n\alpha \frac{v - \frac{1}{\alpha}}{v-1} = n\alpha \frac{v\alpha - 1}{v\alpha - \alpha} = n\frac{v\alpha - 1}{v - 1}
$$

\begin{table*}
\centering
\begin{tabular}{lllllll}
\hline
\textit{} & \textbf{64k} & \textbf{128k}  & \textbf{500k} & \textbf{1M} & \textbf{10M} & \textbf{50M}\\
\hline
\hline
\textit{QA1 -- SINGLE SUPPORTING FACT} &  &   &  &  &  & \\
\hline
GPT-4 (Few-shot) & 30.0 & 24.0 & - & - & - & - \\
GPT-4 + RAG (Few-shot) & 50.0 & 56.0 & 50.0 & 56.0 & 16.0 & - \\
RMT (137M) & 99.6 & 99.1 & 96.4 & 94.2 & 76.4 & - (64,8*)\\
RMT-R & 99.7 & 99.5 & 97.5 & 97.4 & 86.0 & - \\
Mamba (130M) & \textbf{100\tiny$\pm$0.0}  & 99.5\tiny$\pm$0.2&  92.3\tiny$\pm$1.1 &  - &  - & -\\
\textbf{ARMT (145M)} & \textbf{100\tiny$\pm$0.0}  & \textbf{99.9\tiny$\pm$0.2}&  \textbf{99.3\tiny$\pm$0.9} &  \textbf{98.5\tiny$\pm$1.0} & 89.4\tiny$\pm$8.1 & 49,6\tiny$\pm$40.4 \normalsize(79,9*)\\
\hline
\hline
\textit{QA2 -- TWO SUPPORTING FACTS} &  &  &  &  & \\
\hline
GPT-4 (Few-shot) & 4.0 & 8.0 & - & - & - & -\\
RMT (137M)& 72.7 & 56.3 & 32.0 & 25.5 & 16.2 & -\\
RMT-R & 71,6 & 54,9 & 31.8 & 26.3 & 13.0 & -\\Mamba (130M) & 95.0\tiny$\pm$4.2  & 86.7\tiny$\pm$6.2 & - &  - &  - & - \\
\textbf{ARMT (145M)}  & \textbf{100\tiny$\pm$0.0} & \textbf{99.8\tiny$\pm$0.2} & \textbf{99.4\tiny$\pm$0.3} & \textbf{99.4\tiny$\pm$0.3} &   \textbf{84.4\tiny$\pm$4.0} & -\\
\hline
\hline
\textit{QA3 -- THREE SUPPORTING FACTS} &  &  &  &  & \\
\hline
GPT-4 (Few-shot) & 12.0 & 4.0 & - & - & - & - \\
RMT (137M) & 51.9 & 42.9 & 25.9 & 24.8 & 21.0 & -\\
RMT-R & 52.9 & 41.9 & 25.5 & 22.2 & 16.4 & - \\Mamba (130M) & \textbf{91.8\tiny$\pm$0.3}  & \textbf{81.4\tiny$\pm$1.0}& - &  - &  - & - \\
\textbf{ARMT (145M)}  & \textbf{90.4\tiny$\pm$2.2} & \textbf{86.0\tiny$\pm$4.8} & \textbf{79.7\tiny$\pm$10.8} & \textbf{72.1\tiny$\pm$14.2} &   \textbf{37.0\tiny$\pm$10.3} & - \\
\hline
\hline
\textit{QA4 -- TWO ARG RELATIONS} &  &  &  &  & \\
\hline
GPT-4 (Few-shot) & 20.0 & 36.0 & - & - & - & - \\
RMT (137M) & 51.2 & 40.0 & 29.4 & 27.3 & 17.2 & -\\
RMT-R & 58.8 & 50.1 & 32.1 & 26.0 & 14.0 & - \\Mamba (130M) & 99.7\tiny$\pm$0.2  & \textbf{97.6\tiny$\pm$2.8}& - &  - &  - & - \\
\textbf{ARMT (145M)}  & \textbf{100\tiny$\pm$0.1} & \textbf{100\tiny$\pm$0.1} & \textbf{99.9\tiny$\pm$0.2} & \textbf{99.8\tiny$\pm$0.3} &   \textbf{91.5\tiny$\pm$1.7} & - \\
\hline
\hline
\textit{QA5 -- THREE ARG RELATIONS} &  &  &  &  & \\
\hline
GPT-4 (Few-shot) & 64.0 & 48.0 & - & - & - & - \\
RMT (137M) & 88.5 & 78.1 & 56.4 & 48.0 & 27.3 & -\\
RMT-R & 86.2 & 77.4 & 55.9 & 49.9 & 35.0 & - \\Mamba (130M) & \textbf{98.7\tiny$\pm$0.1}  & \textbf{97.5\tiny$\pm$1.1}& - &  - &  - & - \\
\textbf{ARMT (145M)}  & \textbf{99.0\tiny$\pm$0.3} & \textbf{98.7\tiny$\pm$0.4} & \textbf{98.4\tiny$\pm$0.3} & \textbf{97.3\tiny$\pm$0.6} &   \textbf{80.9\tiny$\pm$7.6} & - \\
\hline
\end{tabular}
\caption{\label{tab:babilong}
Exact match metric on QA1-5 Babilong subsets. Each column corresponds to some constant context length. Context includes both noise sentences and facts. * The 50M exact-match on QA1 is measured on 1 best model. ARMT rows are 3 runs averaged. Mamba rows are 2 runs averaged. The metric is marked bold if its $\pm$std interval intersects the $\pm$std interval of the best model.
}
\end{table*}

\begin{table}[]
    \centering
    \begin{tabular}{|c|c|c|c|c|c|}
    \hline
            & Segmentation & Memory Capacity & Working Memory & LM & Length extrapolation \\
    \hline
    Mamba   & \XSolidBrush & \CheckmarkBold & \CheckmarkBold & \CheckmarkBold \CheckmarkBold & \CheckmarkBold \\
    \hline
    RMT     & \CheckmarkBold & ? & \CheckmarkBold  & ? & \CheckmarkBold \\
    \hline 
    ARMT    & \CheckmarkBold & \CheckmarkBold\CheckmarkBold & \CheckmarkBold & ? & \CheckmarkBold \CheckmarkBold \\
    \hline
    \end{tabular}
    \caption{Models abilities.}
    \label{tab:abilities}
\end{table}
\section{Curriculum learning}
\label{sec:curriculum}
We train all models with curriculum learning. This means we incrementally increase the complexity of the task during the training. In particular, we train all models on short sequences first and then increase the length of the sequences until it reaches the maximum length (16k tokens for babilong experiments, 200 pairs for Associative Retrieval Remember, 50 pairs for Associative Retireval Rewrite, and 1024 tokens for language modeling experiments (8 segments, 128 each)).

\section{Babilong training details}
\label{sec:babilong_appx}
We consider segments of size 512 for RMT and ARMT to process the long sequences. The curriculum learning process uses the following number of sequences consecutively: 2, 3, 5, 8, 16, 32. So the training ends when we finish training on 32 segments, 512 tokens each. We also randomly sample the number of segments during training, as we find it helps the model generalize better.

\section{Associative Retrieval training details}
\label{sec:ar_training_appx}
Due to the task's simplicity and training efficiency, we are considering small models (about 500k parameters each) for the Associative Retrieval dataset studies. Every model that we compare has four layers. 128 is the hidden dimension. If the memory dimension parameter (state size in Mamba and memory dimension in ARMT) is present in the model, it is assumed to be 32; if the contrary is not indicated. 

Moreover, if the model supports segmentation (like RMT and ARMT), we use different segments for different key-value pairs. Thus, if we have, for instance, 200 pairs, 200 segments are passed through the model, and after that, in the 201st segment, we expect the model to generate the value. Both keys and values consist of several integers from 0 to 15.

We also use the curriculum with the following number of key-value pairs: 1, 2, 3, 5, 10, 20, 40, 50, and 200. We increase the key size if necessary, so the final key size for remember task is 3 (so we have $16^3$ unique keys) and for rewrite task it remains 1 (16 unique keys). For the \textbf{Remember task}, we also consider sampling different numbers of pairs during training for better generalization.

\begin{figure}

\subfigure[]{
    \includegraphics[width=0.56\textwidth]{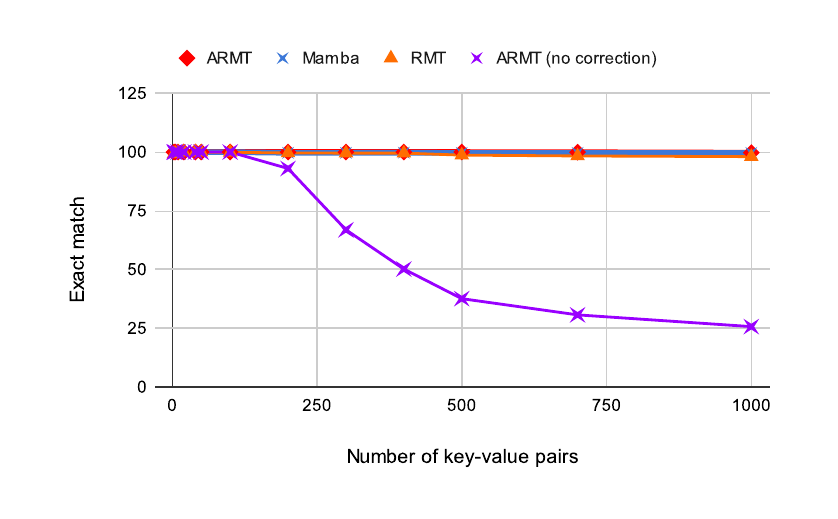}
    \label{fig:rewrite_ablation}
}
\hspace{-15pt}
\subfigure[]{
    \includegraphics[scale=0.64]{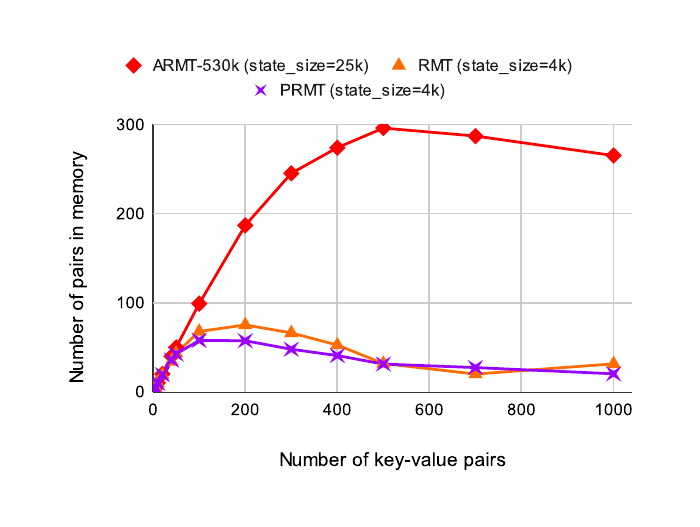}
    \label{fig:prmt}
}
\caption{(a) \textbf{$\gamma$-correction cures the quasi-linear attention memory.} Without correction, the quasi-linear attention with delta-rule struggles to extrapolate on unseen amounts of memory updates. (b) \textbf{Parallel memory doesn't solve the capacity issue.} This means that the associative memory plays an important role in increasing the capacity of the memory.}
\end{figure}
\section{Ablation}
\label{app:ablation}

\subsection{Gamma-correction}
\label{sec:gamma_appx}
Due to an improper normalization vector $z$ update, the proposed fast-weights technique \citep{schlag2021linear} (also known as delta-rule) does not have the length generalization (\cref{fig:rewrite_ablation}). The information in the association matrix $A$ is erased, but not from $z$, which is the source of the issue.

Note that $z$ is the sum of $\phi(k_i)$. Moreover, we recall the information from previous segments using the inner product of $\phi(k_i)$ and $\phi(q_i)$. This means that for accurate recall, all $\phi(k_i)$ should be orthogonal to each other. Therefore, we can expect $z$ to be a sum of approximately orthogonal vectors.

In this sense, we simplify our task to the task of removing the $\phi(k_i)$ from the sum of vectors orthogonal to $\phi(k_i)$, with the exception of the very $\phi(k_i)$. This means that the presence of $\phi(k_i)$ can be measured by computing the inner product between this sum and $\phi(k_i)$ and dividing it by the length of $\phi(k_i)$ (just taking an orthogonal basis component).

After the insertion of the new information into our memory, we expect this sum to include only one $\phi(k_i)$. Therefore, our $\gamma$-coefficient can be computed with the following formula:

$$ \gamma_i = 1 - \frac{(z_{s-1})^T\phi(k_i)}{\|\phi(k_i)\|^2};$$

$$ z_s = z_{s-1} + \sum_i \gamma_i \phi(k_i).$$

The inner product $(z_{s-1})^T\phi(k_i)$ is divided by the square of $\|\phi(k_i)\|$ because the gamma will be further multiplied by $\phi(k_i)$.

We also consider detaching the gamma during training, because it seems to converge better in this case.

\begin{figure*}
\centering
\vspace{-15pt}
\includegraphics[width=0.7\textwidth]{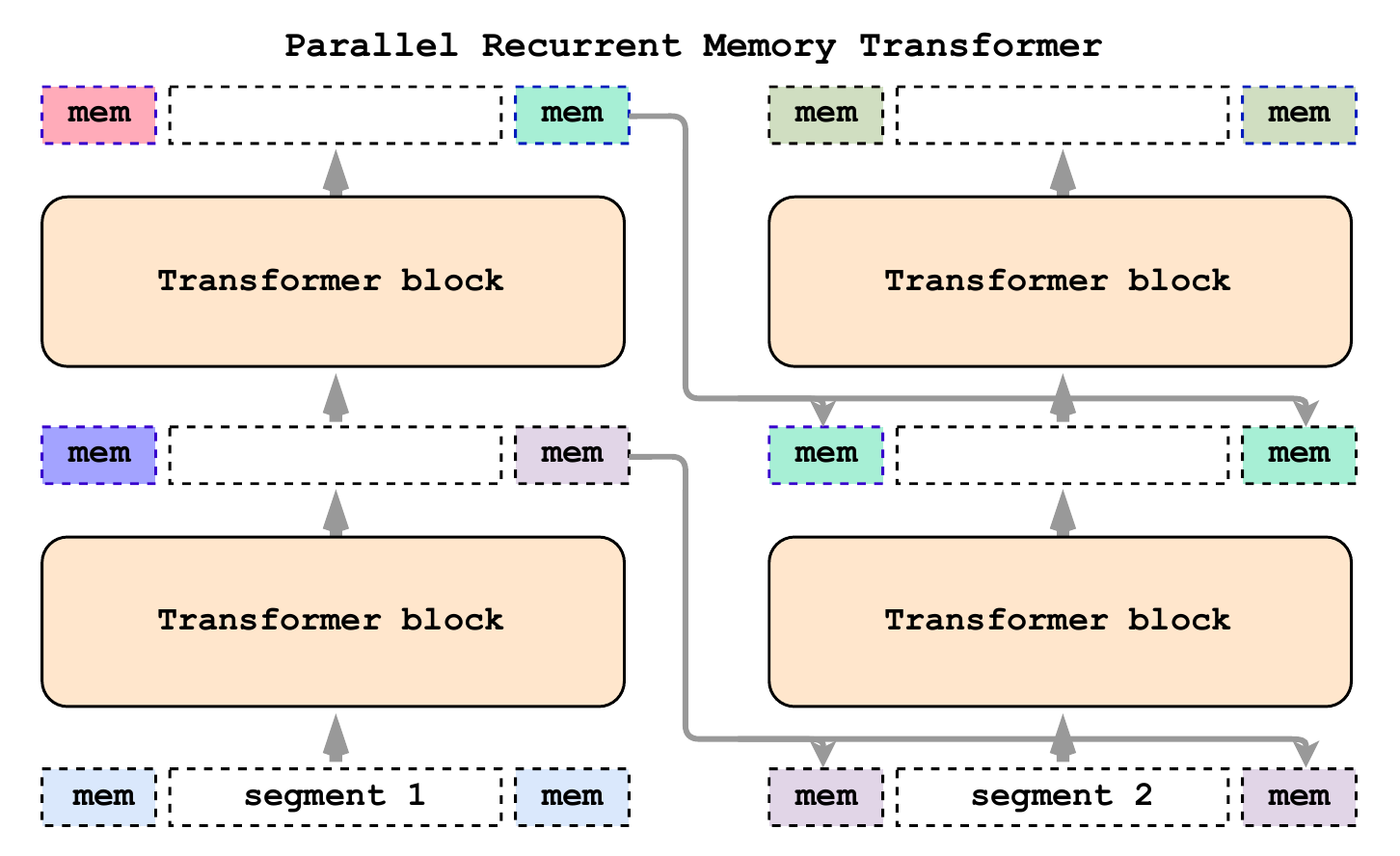}
\caption{\textbf{Parallel recurrent memory transformer.} In contrast to RMT, in PRMT memory tokens are passed to the next segment in each layer.}
\label{fig:prmt_fig}
\vspace{-10pt}
\end{figure*}

\subsection{Associative memory ablation}
\label{sec:prmt}

To understand, if the associative memory important for memorization tasks, we consider another architecture: Parallel-memory RMT (PRMT) \cref{fig:prmt_fig}

It is different from RMT in the hierarchical memory approach, considering the memory shifts layerwise, just like in ARMT. So this architecture can be thought of as RMT with layerwise memory, while ARMT is RMT with layerwise memory organized in association matrices.

\begin{figure}
    \centering
    \includegraphics[scale=0.8]{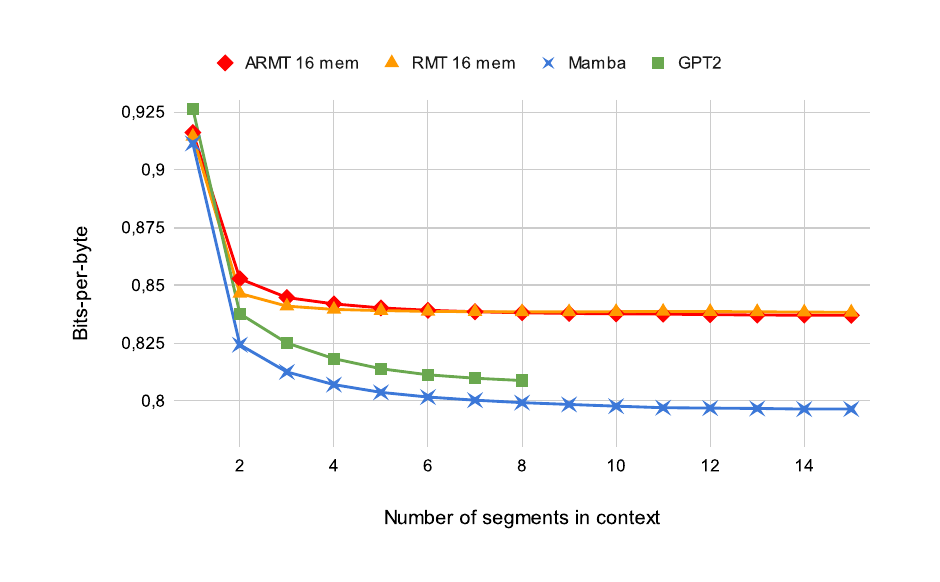}
    \caption{\textbf{ARMT performs similarly to RMT on the language modeling task.} Wikitext-103 results. The loss on each of the 128-sized segments on the test dataset (Normalized with Bits-per-byte \citep{gao2020pile}). The model is trained for a language modeling task on 8 segments, 128 tokens each. Despite the larger estimated capacity, ARMT struggles to solve the LM task well.}
    \label{fig:lm}
\end{figure}

\section{Language Modeling experiments}
\label{sec:lm_appx}
We utilized the Wikitext-103 dataset to train ARMT and RMT models in order to assess our architecture's performance on real texts. Next, we examined the cross-entropy losses derived from various model segments on the test dataset, as illustrated in Figure \ref{fig:lm}. In this manner, we can estimate the amount of language data that can be stored in memory.

Nevertheless, we demonstrate that despite having a larger theoretical capacity than RMT, ARMT still performs similarly to RMT in language modeling.

We use the GPT-2 model as the base model for our architecture changes. We consider the RMT and ARMT models' segment sizes to be equal to 128 tokens and train these models to solve the language modeling task on 8 segments, so in total, we train the model to autoregressively predict 1024 tokens. Then we evaluate the models' performance on each of the 15 segments of test texts (1920 tokens).
\section{Why is mamba slow for long contexts?}
\label{sec:mamba_slow}
\begin{figure}
    \centering
    \includegraphics[scale=0.7]{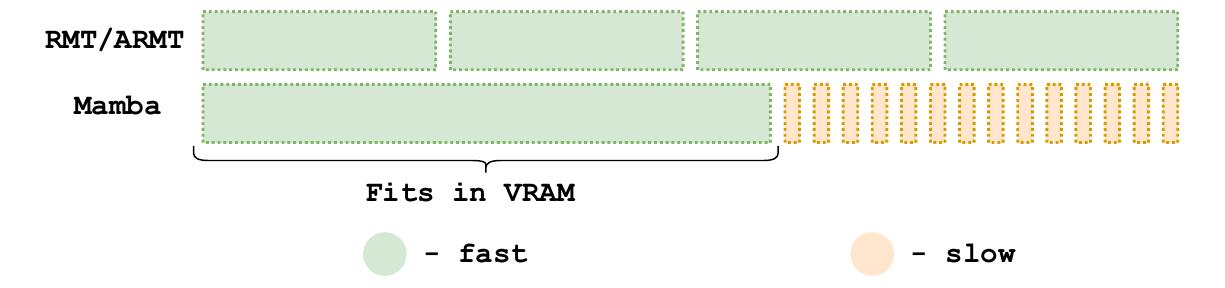}
    \caption{\textbf{Current mamba implementation allows only the first segment to be long.} Other tokens have to be processed consecutively one by one.}
    \label{fig:segmentation}
\end{figure}

We faced some difficulties in evaluating mamba on long-context (500k+) due to it's specific segmentation abilities shown in Figure \ref{fig:segmentation}.

\section{Associative Retrieval sample structure}

A sample of \textbf{Remember} dataset contains a concatenated context, query, and answer. The context is a set of key-value pairs $(k, v)$, separated by a special token. All keys are sequences of tokens. Tokens in this sequence can intersect, but the whole sequence corresponding to any key is unique in this particular sample. The query is one of the keys in the context. And the answer is a value corresponding to the key from the query. Thus, we can control the number of pairs in the sample and check how many pairs fit in our memory.

This is how the dataset's sample appears:
\begin{verbatim} 
<key1>:<value1>,<key2>:<value2>, <key3>:<value3>,<key2>-<value2>
\end{verbatim}
The model is thought to be trained to produce the value following the "\texttt{-}" character.



\section{RWKV-5 model}
\label{app:rwkv}
We also tried to train the RWKV-v5 \citep{peng2023rwkv} model to slove both associative retrieval and babilong tasks (training from scratch for AR tasks and finetuning 430M model for babilong task). Unfortunately, the model was training poorly and hadn't achieved reasonable scores. We used the very same parameters as for training other models. Perhaps the RWKV training process requires accurate adjustments. However, we haven't succeeded.

\section{Limitations}
\label{sec:limit}
The proposed architecture, however, has several drawbacks. The first is the lack of efficient parallel implementation: you have to process all segments consecutively. This doesn't mean that it is slow. It's fast enough to process millions of tokens in a reasonable time. However, on short and medium-length sequences (less than 300k tokens), it is much slower than, for instance, Mamba and RWKV, which have the parallel form. Moreover, as we have shown in \cref{sec:lm_appx}, it's still challenging to train ARMT to solve the language modeling task well. However, we believe that this problem is not in the very architecture, but in training process: we observe that ARMT tends to keep in memory only the last segment, and therefore struggles to extrapolate on longer sequences.
\end{document}